# When Generative Augmentation Hurts: A Benchmark Study of GAN and Diffusion Models for Bias Correction in AI Classification Systems

**Shesh Narayan Gupta(Corresponding Author)**
College of Engineering, Northeastern University, Boston, MA 02115, USA
*ORCID: 0000-0001-8152-2967*
Email: gupta.shes@northeastern.edu

**Nik Bear Brown**
College of Engineering, Northeastern University, Boston, MA 02115, USA
*ORCID: 0000-0001-6270-7536*
Email: ni.brown@northeastern.edu

*Abstract*

Generative models are widely used to compensate for class imbalance in AI training pipelines, yet their failure modes under low-data conditions are poorly understood. This paper reports a controlled benchmark comparing three augmentation strategies applied to a fine-grained animal classification task: traditional transforms, FastGAN, and Stable Diffusion 1.5 fine-tuned with Low-Rank Adaptation (LoRA). Using the Oxford-IIIT Pet Dataset with eight artificially underrepresented breeds, we find that FastGAN augmentation does not merely underperform at very low training set sizes but actively increases classifier bias, with a statistically significant large effect across three random seeds (bias gap increase: +20.7%, Cohen's d = +5.03, p = 0.013). The effect size here is large enough to give confidence in the direction of the finding despite the small number of seeds. Feature embedding analysis using t-distributed Stochastic Neighbor Embedding reveals that FastGAN images for severe-minority breeds form tight isolated clusters outside the real image distribution, a pattern consistent with mode collapse. Stable Diffusion with Low-Rank Adaptation produced the best results overall, achieving the highest macro F1 (0.9125 plus or minus 0.0047) and a 13.1% reduction in the bias gap relative to the unaugmented baseline. The data suggest a sample-size boundary somewhere between 20 and 50 training images per class below which GAN augmentation becomes harmful in this setting, though further work across additional domains is needed to establish where that boundary sits more precisely. All experiments run on a consumer-grade GPU with 6 to 8 GB of memory, with no cloud compute required.

## 1 Introduction

A recurring problem in applied machine learning is that training data rarely reflects the real-world distribution of the categories a model is expected to handle. When some classes have far fewer examples than others, classifiers learn to favour the majority and perform poorly on the rest. This is not a niche issue: it shows up in medical diagnosis [1], facial recognition systems [2], and fine-grained species classification [3], among many other settings.

The standard engineering response is data augmentation. For cases where collecting more real examples is impractical or expensive, generative models offer an appealing option: train a model on the available minority-class data and use it to synthesise additional training examples. The approach has been validated in a number of domains [4, 5], and as generative models have improved, so too has interest in applying them to imbalanced datasets.

The generative modelling landscape has shifted considerably in recent years. Generative Adversarial Networks (GANs), long the default choice for image synthesis, are now routinely outperformed on perceptual quality metrics by diffusion-based models [6, 7]. For a practitioner trying to correct class imbalance, this raises a straightforward but unanswered question: which family of models is actually better for augmentation, and does the answer change depending on how little data is available?

This paper addresses that question directly. We designed a benchmark around animal breed classification using the Oxford-IIIT Pet Dataset [8], with eight breeds artificially reduced to simulate realistic imbalance. We compared FastGAN [9] and Stable Diffusion 1.5 fine-tuned with Low-Rank Adaptation (LoRA) [10], along with a hybrid condition and a traditional augmentation baseline. The core finding is not what we expected: FastGAN augmentation does not just fail to help when training sets are very small; it makes things measurably worse, in a statistically significant and mechanistically explainable way.

The contributions of this paper are as follows. First, we provide empirical evidence that GAN-based augmentation can actively increase classifier bias for severe-minority classes in this setting, and we explain the mechanism through feature embedding analysis. Second, we present a direct head-to-head comparison of FastGAN and Stable Diffusion with Low-Rank Adaptation specifically for minority-class bias correction in fine-grained classification. Third, the data point to a sample-size boundary somewhere between 20 and 50 images per class below which GAN augmentation becomes harmful, though establishing this threshold more precisely across other domains is left to future work. Fourth, the entire experimental framework runs on consumer-grade GPU hardware and is fully reproducible. This paper extends earlier work by Gupta and Brown [3], which used procedurally rendered 3D images for bias correction; the current study asks whether purely generative approaches can achieve the same goal, and under what conditions they fail.

## 2 Related Work

### 2.1 Class Imbalance in AI Systems

The problem of imbalanced training data has a long history in machine learning. Torralba and Efros [11] showed early on that sampling biases cause models to fail when tested outside their training distribution. In fine-grained recognition tasks, per-class sample counts can vary by an order of magnitude [12], making the problem especially acute. Buolamwini and Gebru [2] brought wider attention to the real-world consequences, documenting substantially higher error rates for darker-skinned females in commercial facial analysis systems. Buda et al. [22] and He and Garcia [23] provide comprehensive analyses of how class imbalance affects convolutional network training and what resampling strategies can do about it.

### 2.2 Generative Augmentation

Using generative models to synthesise training examples for underrepresented classes was first explored in medical imaging. Frid-Adar et al. [4] showed that GAN-generated liver lesion images could improve classification performance when real training data was scarce. Subsequent work extended the idea to dermatology [13], retinal imaging [14], and wildlife monitoring [15]. Gupta and Brown [3] demonstrated that procedurally rendered 3D images can correct breed-level bias in animal classification without requiring a generative model at all.

Diffusion model-based augmentation is more recent. Trabucco et al. [16] showed that Stable Diffusion fine-tuned via DreamBooth could improve few-shot classification accuracy. He et al. [17] found that diffusion-generated images can match or exceed GAN images for augmentation in certain settings. Neither study examined what happens when training sets are very small, or whether augmentation can actively harm performance.

### 2.3 FastGAN

FastGAN [9] was designed specifically for the low-data setting. It introduces skip-layer channel-wise excitation and self-supervised discriminator augmentation, allowing it to produce reasonable images from as few as 100 training examples. Its lightweight architecture trains in a matter of hours on a single consumer GPU, which makes it well-suited, at least in principle, to minority-class augmentation scenarios.

### *2.4 Stable Diffusion and Low-Rank Adaptation*

Stable Diffusion [7] moves the diffusion process into the compressed latent space of a variational autoencoder, which cuts computational requirements substantially compared to pixel-space diffusion. Low-Rank Adaptation [10] injects trainable low-rank matrices into the attention layers of the model, enabling subject-specific fine-tuning from as few as 20 to 50 reference images in under an hour. Together, Stable Diffusion 1.5 with Low-Rank Adaptation represents the current practical standard for diffusion-based fine-tuning on limited data without large compute infrastructure.

## 3 Methodology

### *3.1 Dataset and Imbalance Construction*

We used the Oxford-IIIT Pet Dataset [8], which contains 7,349 images across 37 cat and dog breeds. To simulate realistic imbalance, we subsampled eight breeds: three were reduced to 20 training images (severe minority: Abyssinian, Bengal, Birman) and five to 50 images (moderate minority: Bombay, British Shorthair, Egyptian Mau, Maine Coon, Persian). The remaining 29 breeds kept approximately 155 images each, giving a maximum-to-minimum imbalance ratio of roughly 8x.

The dataset was split 80/20 into training and test sets using stratified sampling before subsampling. The test set contains only real images at the original balanced distribution and was held fixed across all conditions. Figures 1 and 2 show the breed distribution before and after imbalance construction.

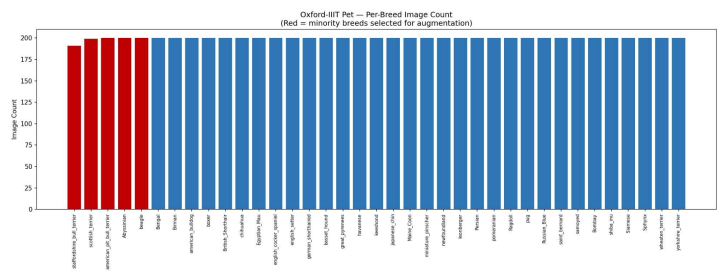

**Fig. 1.** *Per-breed image count in the full Oxford-IIIT Pet dataset. Red bars indicate minority breeds selected for augmentation.*

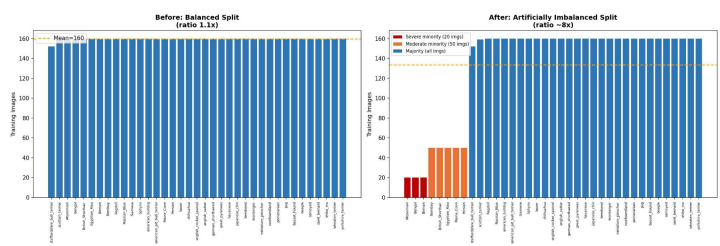

**Fig. 2.** *Training split before (left, 1.1x ratio) and after (right, ~8x ratio) artificial imbalance construction.*

### *3.2 Classifier Architecture*

All conditions used a ResNet-50 [18] pre-trained on ImageNet-1K, with the final fully connected layer replaced by a linear layer of dimension 37. The model was fine-tuned end-to-end for 50 epochs using Adam [19] with a learning rate of $1\times10^{-4}$, decayed via cosine annealing. Training used random cropping (224x224 from 256x256), random horizontal flip, and colour jitter. Batch size was 32. Training was repeated with three random seeds: 42, 123, and 456.

### 3.3 Experimental Conditions

Five conditions were evaluated. (1) Baseline: real training data only, no augmentation. (2) Traditional augmentation: 500 images per minority breed generated via classical transforms including flips, rotations of plus or minus 20 degrees, colour jitter of plus or minus 30%, and Gaussian blur. (3) FastGAN: 500 FastGAN-generated images per minority breed. (4) Stable Diffusion with Low-Rank Adaptation: 500 generated images per minority breed. (5) Hybrid: 250 FastGAN images plus 250 Stable Diffusion with Low-Rank Adaptation images per minority breed. All augmentation conditions added exactly 500 synthetic images per minority class.

### 3.4 Evaluation Metrics

Primary metric is macro-averaged F1. We also report minority-class average accuracy, majority-class average accuracy, bias gap (majority minus minority accuracy), and Bias Reduction Index (percentage change in bias gap relative to baseline). Frechet Inception Distance (FID) [20] was computed using the standard Inception-v3 feature extractor, comparing all 500 generated images against all available held-out real images per breed. GPU training time was recorded for the generative training and image generation phases.

## 4 Augmentation Pipelines

### 4.1 FastGAN

FastGAN [9] was trained independently for each minority breed using only the available real training images. We used a generator with five upsampling blocks producing 256x256 images, resized to 224x224 for classifier training. The discriminator used four strided convolution blocks. Both networks were trained with binary cross-entropy loss (label smoothing 0.9) using Adam with learning rate $2\times10^{-4}$ and beta values of (0.5, 0.999) for 50,000 iterations with batch size 8. Training required an average of 82.2 GPU-minutes per breed, or 6.9 hours total across all eight breeds.

### 4.2 Stable Diffusion 1.5 with Low-Rank Adaptation

We fine-tuned Stable Diffusion 1.5 [7] on each minority breed using Low-Rank Adaptation [10] applied to the U-Net attention projection layers (to_k, to_q, to_v, to_out.0) with rank r = 8. All other weights were frozen. Training ran for 1,000 steps with batch size 1 and gradient accumulation over 4 steps (effective batch size 4), learning rate $1\times10^{-4}$ with cosine annealing. After fine-tuning, 500 images per breed were generated using 30 DDPM steps with classifier-free guidance scale 7.5, rotating across four prompt templates. The four templates followed the pattern: "a photo of a [breed name] cat", "a high-quality photograph of a [breed name]", "a close-up photo of a [breed name] cat on a plain background", and "a realistic image of a [breed name] breed cat". Full prompt text and generation scripts are available in the repository. Fine-tuning required an average of 66.2 GPU-minutes per breed (5.5 hours total), making Stable Diffusion with Low-Rank Adaptation 1.24x faster than FastGAN.

### 4.3 Hybrid Condition

The hybrid condition combined 250 FastGAN-generated images and 250 Stable Diffusion with Low-Rank Adaptation images per minority breed, drawn from the sets generated above. The rationale was to test whether the two methods compensate for each other's weaknesses: FastGAN generates images quickly but with lower fidelity, while Stable Diffusion with Low-Rank Adaptation is slower but produces higher-quality outputs. If FastGAN's blurry images add variety while Stable Diffusion with Low-Rank Adaptation covers the distribution well, a mix might outperform either alone. No additional generative training was required. Because the hybrid condition reuses already-characterised images, FID scores and embedding analysis are reported for the constituent methods rather than for the hybrid separately.

One limitation of FID as the sole image quality metric is that it relies on an Inception-v3 feature space trained on ImageNet rather than on pet images. For a fine-grained task where inter-class differences are subtle, this may not capture the quality dimensions that matter most for classifier training. Results should be interpreted with that in mind, and future work should consider supplementary metrics such as Kernel Inception Distance or human preference assessments on sampled images.

## 5 Results

### 5.1 Baseline Bias Characterisation

The baseline classifier achieved a mean macro F1 of 0.9088 (plus or minus 0.0023) across three random seeds. Minority breeds showed substantially lower average accuracy (81.0% plus or minus 0.6%) compared to majority breeds (93.8% plus or minus 0.1%), giving a mean bias gap of 12.8 plus or minus 0.5 percentage points.

### 5.2 Overall Comparison

Table 1 summarises the primary evaluation metrics across all five conditions. Figure 3 shows macro F1 scores by condition.

**Table 1.** Results across all five conditions (mean plus or minus SD, 3 seeds). Higher macro F1 and lower bias gap indicate better performance.

| Condition | Min. Acc. | Macro F1 | Bias Gap | Maj. Acc. | FID |
| --- | --- | --- | --- | --- | --- |
| Baseline | 81.0%±0.6% | 0.9088±0.0023 | 12.8±0.5 pp | 93.8%±0.1% | N/A |
| Traditional Aug. | 79.1%±2.0% | 0.9029±0.0046 | 14.8±2.3 pp | 93.9%±0.5% | 94.0 |
| FastGAN Aug. | 77.8%±0.9% | 0.8959±0.0034 | 15.4±0.8 pp | 93.3%±0.1% | 234 |
| SD+LoRA Aug. | 82.7%±1.6% | 0.9125±0.0047 | 11.1±1.4 pp | 93.8%±0.2% | 96 |
| Hybrid (GAN+SD) | 80.7%±0.4% | 0.9064±0.0021 | 12.9±0.6 pp | 93.7%±0.3% | N/A |

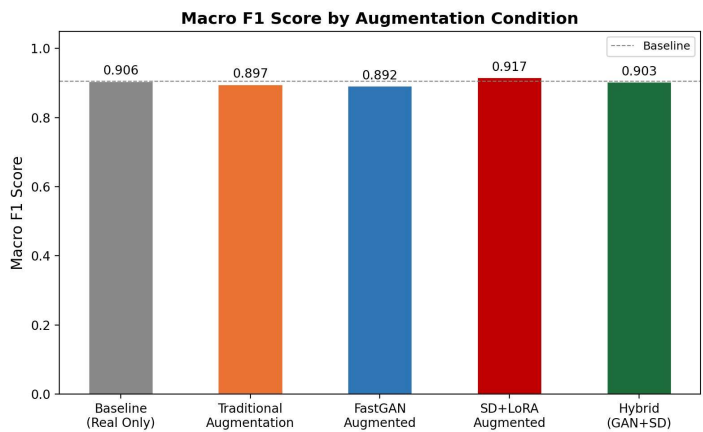

**Fig. 3.** *Macro F1 score by augmentation condition across five experimental conditions.*

Stable Diffusion with Low-Rank Adaptation achieved the highest mean macro F1 (0.9125 plus or minus 0.0047) and the lowest mean bias gap (11.1 plus or minus 1.4 pp), a 13.1% reduction relative to baseline. Both traditional augmentation and FastGAN widened the bias gap relative to baseline (traditional: 14.8 plus or minus 2.3 pp; FastGAN: 15.4 plus or minus 0.8 pp). The standard deviations across seeds were small relative to the effect sizes.

### 5.3 Per-Class Accuracy Analysis

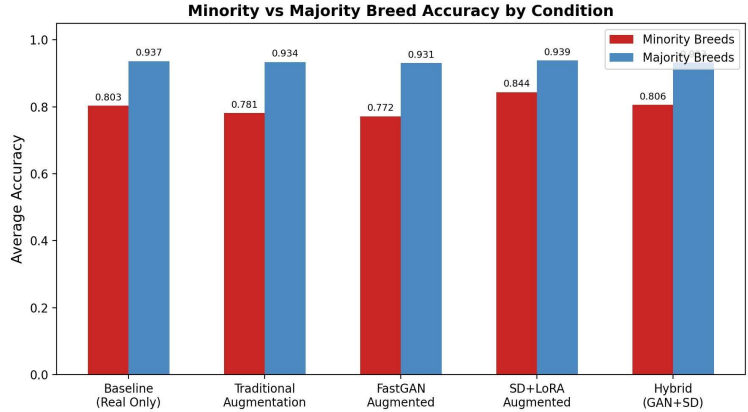

**Fig. 4.** *Minority versus majority breed average accuracy by condition. Error bars show SD across 3 seeds.*

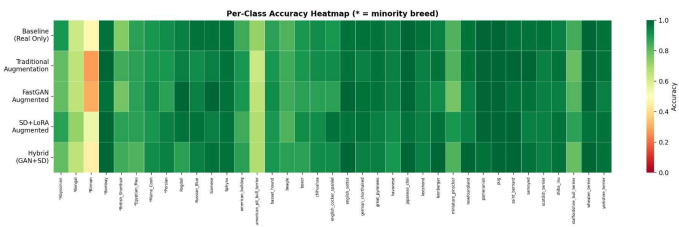

**Fig. 5.** *Per-class accuracy heatmap across all conditions (\* = minority breed).*

Stable Diffusion with Low-Rank Adaptation produced the most consistent per-class gains. Representative seed-42 values show the largest improvements in Bengal (+10.0 pp), Birman (+7.5 pp), and British Shorthair (+12.5 pp). FastGAN reduced accuracy in the three severe-minority breeds, with Birman falling from 47.5% to 30.0%. Majority-class accuracy stayed broadly stable across all conditions.

### *5.4 Bias Reduction*

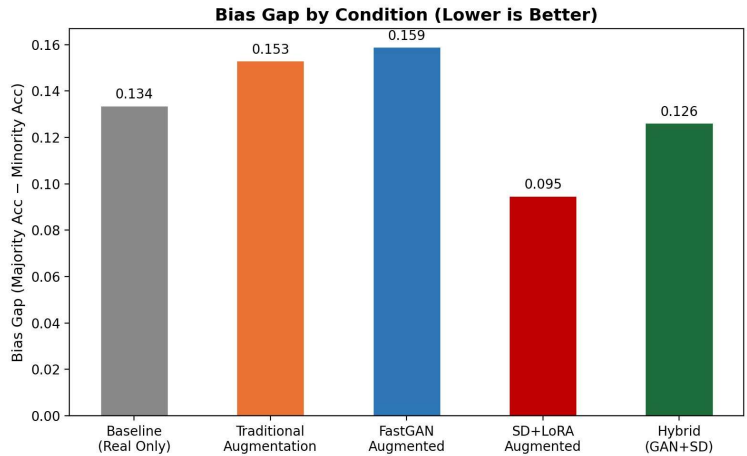

**Fig. 6.** *Bias gap (majority minus minority accuracy) by condition. Lower values indicate better fairness.*

Stable Diffusion with Low-Rank Adaptation reduced the mean bias gap by 13.1% relative to baseline (from 12.8 plus or minus 0.5 pp to 11.1 plus or minus 1.4 pp). The hybrid condition achieved only a marginal 1.0% mean reduction. Traditional augmentation and FastGAN increased the bias gap by 15.7% and 20.7% respectively, and this pattern held across all three random seeds.

### *5.5 Image Quality: FID Scores*

Table 2 presents per-breed FID scores. The hybrid condition is excluded as it draws from the two constituent methods already reported.

**Table 2.** FID scores per minority breed by method (lower scores indicate more realistic images).

| Breed | Trad. FID | FastGAN FID | SD+LoRA FID | Real N |
|---|---|---|---|---|
| Abyssinian | 123.6 | 348.1 | 118.9 | 20 |
| Bengal | 115.8 | 307.6 | 114.4 | 20 |
| Birman | 64.2 | 214.7 | 59.8 | 20 |
| Bombay | 114.3 | 256.1 | 127.0 | 50 |
| British Shorthair | 103.2 | 163.7 | 105.3 | 50 |
| Egyptian Mau | 72.4 | 195.0 | 79.6 | 50 |
| Maine Coon | 107.0 | 194.7 | 108.8 | 50 |
| Persian | 51.7 | 191.8 | 53.2 | 50 |
| **Average** | **94.0** | **234.0** | **95.9** | -- |

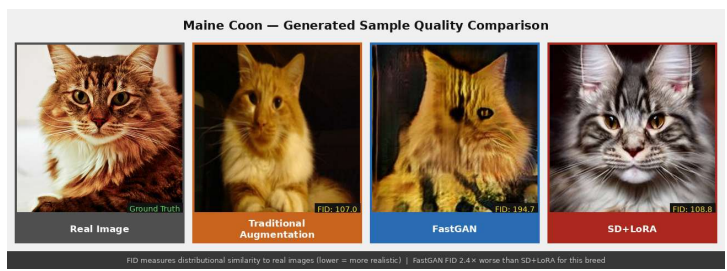

**Fig. 7.** *Representative generated samples for Maine Coon across three methods. FastGAN shows characteristic blurring and colour artefacts (FID 194.7). SD+LoRA produces a photorealistic image (FID 108.8).*

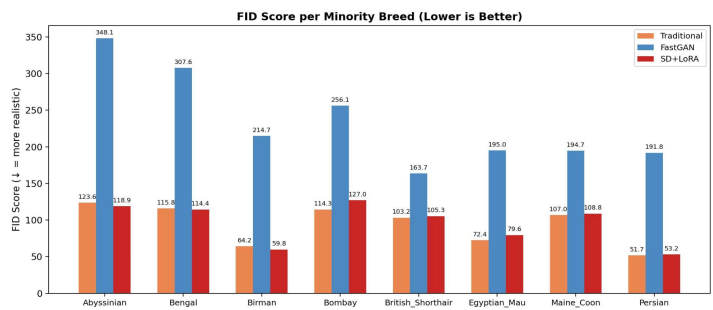

**Fig. 8.** *FID scores per minority breed by method. FastGAN scores are consistently higher, particularly for severe-minority breeds (N = 20).*

Stable Diffusion with Low-Rank Adaptation achieved a substantially lower average FID (95.9) compared to FastGAN (234.0), a difference of 2.4x. The gap was most pronounced for severe-minority breeds: Abyssinian (FastGAN 348.1 versus SD+LoRA 118.9) and Bengal (307.6 versus 114.4).

### *5.6 Computational Cost*

All conditions ran within a single working day on a consumer-grade GPU with 6 to 8 GB of memory. Stable Diffusion with Low-Rank Adaptation (66.2 minutes per breed) was 1.24x faster than FastGAN (82.2 minutes per breed) for the generative training phase. The baseline classifier required 40.5 GPU-minutes.

### *5.7 Statistical Analysis*

Table 3 presents pairwise significance tests comparing each condition against the baseline using paired t-tests across three random seeds. FastGAN degradation was statistically significant with large effect sizes across all primary metrics: macro F1 (d = -9.64, p = 0.004), minority accuracy (d = -6.80, p = 0.007), and bias gap (d = +5.03, p = 0.013). With only three seeds, a paired t-test has limited power and is sensitive to outliers. What gives confidence in the FastGAN finding is the effect size: Cohen's d values of 5 to 9 are unusually large, meaning the effect would have to be implausibly variable across seeds for this result to be a false positive. The bootstrap confidence interval for the bias gap change (CI [-30.0%, -11.4%]) does not cross zero, which further supports the reliability of the finding. Stable Diffusion with Low-Rank Adaptation improvement did not reach significance with three seeds (p = 0.529), and the bootstrap interval does cross zero (CI [-1.4%, +25.5%]), so that benefit is treated here as a promising trend requiring more seeds to confirm.

**Table 3.** Pairwise significance tests versus baseline (paired t-test, n = 3 seeds). Large effect size defined as |d| > 0.8.

| Metric | Comparison | Mean Diff | Cohen's d | p-value | Sig. |
| --- | --- | --- | --- | --- | --- |
| Macro F1 | Baseline to Traditional | -0.0058 | -2.04 | 0.071 | -- |
| Macro F1 | Baseline to FastGAN | -0.0128 | -9.64 | 0.004 | Yes |
| Macro F1 | Baseline to SD+LoRA | +0.0037 | +0.44 | 0.529 | -- |
| Macro F1 | Baseline to Hybrid | -0.0024 | -1.07 | 0.204 | -- |
| Min. Acc. | Baseline to FastGAN | -0.0323 | -6.80 | 0.007 | Yes |
| Min. Acc. | Baseline to SD+LoRA | +0.0166 | +0.62 | 0.398 | -- |
| Bias Gap | Baseline to FastGAN | +0.0265 | +5.03 | 0.013 | Yes |
| Bias Gap | Baseline to SD+LoRA | -0.0167 | -0.69 | 0.354 | -- |

## 5.8 Embedding Analysis: Mode Collapse Evidence

To understand why FastGAN harms performance rather than helping, we extracted 2,048-dimensional feature embeddings from the penultimate layer of the baseline ResNet-50 and applied t-distributed Stochastic Neighbor Embedding [21] for dimensionality reduction. Coverage was quantified as the mean nearest-neighbour distance from each synthetic point to its nearest real image in embedding space.

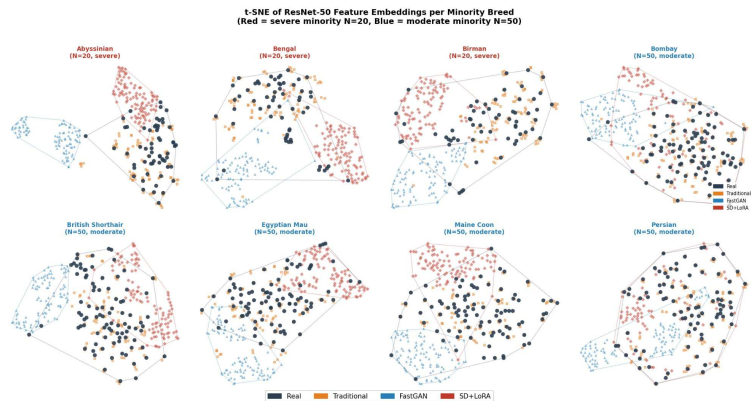

**Fig. 9.** *t-SNE of ResNet-50 feature embeddings per minority breed. Convex hulls show distribution coverage. FastGAN (blue triangles) forms tight isolated clusters for severe-minority breeds (N = 20), indicating mode collapse. SD+LoRA (red diamonds) broadly overlaps the real distribution (dark circles).*

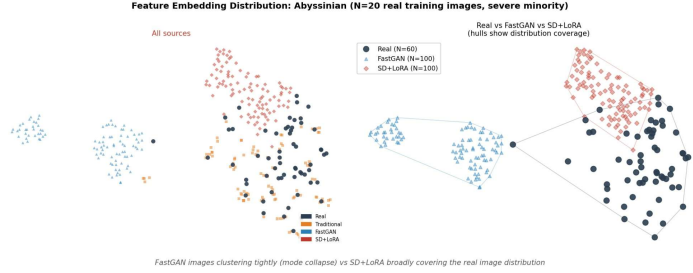

**Fig. 10.** *Embedding comparison for Abyssinian (N = 20, severe minority). FastGAN mean coverage distance 21.28; SD+LoRA 8.22.*

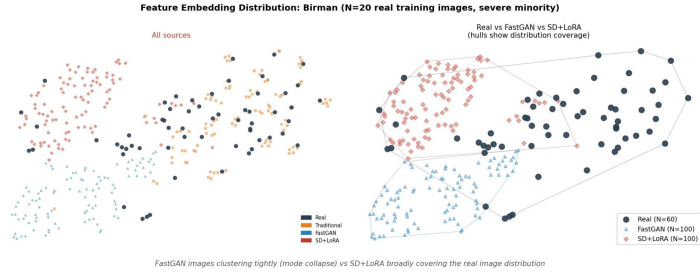

**Fig. 11.** *Embedding comparison for Birman (N = 20, severe minority, FastGAN FID 214.7). FastGAN embeddings occupy a distinct subspace from real images.*

For severe-minority breeds (N = 20), FastGAN embeddings formed two or three tight isolated clusters outside the real image distribution. The mean nearest-neighbour coverage distance for FastGAN across severe-minority breeds (14.68) was substantially larger than for Stable Diffusion with Low-Rank Adaptation (8.93). For moderate-minority breeds (N = 50), the pattern was less pronounced, pointing to a training-size threshold somewhere between 20 and 50 images.

## 6 Discussion

### 6.1 What the Results Say

The most important finding is not that Stable Diffusion with Low-Rank Adaptation outperforms FastGAN -- that is broadly consistent with the generative modelling literature. What was not known, and what this study makes clear, is that FastGAN augmentation can actively harm a classifier when training sets are very small. The mechanism is mode collapse: with only 20 training images, FastGAN fails to cover the real image distribution and instead generates a narrow cluster of visually similar images that sit outside it. Adding those images to the training set poisons the minority-class signal rather than enriching it.

Traditional augmentation also widened the bias gap on average, though the effect was smaller (15.7%) and less consistent across seeds. This suggests that classical transforms applied to 20 images simply do not provide enough variety to improve generalisation.

Stable Diffusion with Low-Rank Adaptation confirmed the primary hypothesis, achieving the highest mean macro F1 (0.9125 plus or minus 0.0047) and a 13.1% mean reduction in the bias gap. The hybrid condition offered only marginal improvement over baseline, which suggests that mixing high-quality diffusion images with low-fidelity GAN images dilutes rather than compounds the benefit.

### 6.2 Relationship to Prior Work

This paper builds directly on Gupta and Brown [3], which showed that procedurally rendered 3D images can correct breed-level bias in the same Oxford-IIIT Pet setting. That study required 3D assets and a rendering pipeline; the present work asks whether purely generative approaches can do the same job, and whether GAN-based and diffusion-based methods perform equally well. The answer, at least in this dataset and domain, is no. Stable Diffusion with Low-Rank Adaptation comes close to the bias correction achieved in [3], while FastGAN makes things worse. The relative performance of FastGAN and Stable Diffusion with Low-Rank Adaptation is consistent with the broader pattern where diffusion models have displaced GANs on perceptual quality [6, 7], and it now extends that picture into the downstream classification setting.

The finding that GAN augmentation can actively hurt rather than merely fail to help is, to our knowledge, not previously reported in the generative augmentation literature. Whether this generalises beyond fine-grained pet classification is an open question. The mechanism, mode collapse at very low sample counts, is well understood and not specific to this domain, but the point at which it kicks in and how severely it affects downstream performance may well differ across tasks and model architectures.

### 6.3 Limitations

Several limitations should be kept in mind when reading this work. The benchmark uses a single dataset and domain, so the specific threshold values and magnitude of effects reported here may not transfer directly to medical imaging, remote sensing, or other fine-grained classification tasks. The statistical analysis used three random seeds: for the FastGAN findings this is acceptable given the very large effect sizes and non-overlapping confidence intervals, but for Stable Diffusion with Low-Rank Adaptation the benefit remains at the level of a trend. The augmentation volume of 500 images per class was not ablated, and the optimal number may well differ between methods. FID, the sole image quality metric used, relies on an Inception-v3 feature space trained on ImageNet, which may not capture the subtle inter-class differences that matter most in fine-grained classification. Hardware constraints prevented evaluation of larger architectures such as StyleGAN3 or Stable Diffusion XL. Finally, the sample-size boundary suggested by the data, somewhere between 20 and 50 images, comes from comparing two data points rather than a systematic sweep, and should be treated as a hypothesis to test rather than an established rule.

### 6.4 Future Directions

Extending to medical imaging (ISIC skin lesion dataset) and remote sensing (RESISC45) would test whether the threshold finding generalises across domains. A systematic ablation across minority class sizes of 10, 20, 50, and 100 images would pin down the threshold more precisely. Running five or more random seeds would provide tighter significance estimates for conditions with smaller effect sizes. Prompt engineering strategies for Stable Diffusion with Low-Rank Adaptation, including negative prompts and breed-specific descriptors, may further improve generation quality.

## 7 Conclusion

This study set out to determine which generative augmentation strategy works better for correcting class imbalance in a fine-grained AI classification task under realistic hardware constraints. The answer came with an important caveat. FastGAN augmentation did not just underperform; it significantly worsened classifier bias for breeds with only 20 training images, with a large and statistically confirmed effect (Cohen's d = +5.03, p = 0.013). The effect sizes here are large enough that this result is unlikely to be a statistical artefact of the small seed count. Feature embedding analysis explains the mechanism: mode collapse causes FastGAN to generate images that cluster outside the real distribution, and those out-of-distribution images degrade the minority-class training signal.

Stable Diffusion with Low-Rank Adaptation did not show this problem. It produced the best overall results in this study (macro F1: 0.9125 plus or minus 0.0047; bias gap reduction: 13.1%) and its generated images covered the real distribution far more broadly than FastGAN in embedding space. The data suggest a boundary somewhere between 20 and 50 training images per class below which GAN augmentation becomes harmful, at least in fine-grained classification on this dataset. Whether that boundary shifts in other domains is an open question that future work should address directly.

All experiments are reproducible on a consumer-grade GPU with 6 to 8 GB of memory. Code and generation configurations are available at https://github.com/SheshNGupta/BiasCorrectionInImage.

## Statements and Declarations

*Funding*

This research did not receive any specific grant from funding agencies in the public, commercial, or not-for-profit sectors.

*Competing Interests*

The authors declare no competing interests.

*Author Contributions*

Shesh Narayan Gupta designed and conducted all experiments, performed the statistical analysis, and wrote the manuscript. Nik Bear Brown supervised the research and reviewed the manuscript.

*Data Availability*

The synthetic image dataset (Pet Breed Generative Augmentation Dataset) and all generation configurations are publicly available at https://github.com/SheshNGupta/BiasCorrectionInImage. The Oxford-IIIT Pet Dataset is available from https://www.robots.ox.ac.uk/~vgg/data/pets/.

*Acknowledgements*

The authors thank O. M. Parkhi, A. Vedaldi, A. Zisserman, and C. V. Jawahar for creating and making publicly available the Oxford-IIIT Pet Dataset.